\documentclass[letterpaper, 10 pt, conference]{ieeeconf}  

\IEEEoverridecommandlockouts                              

\usepackage{algorithm}
\usepackage{algorithmic}
\usepackage{amsmath}
\usepackage{amssymb}
\usepackage{wrapfig}
\usepackage{soul}
\usepackage{hyperref}
\usepackage{newfloat}
\usepackage{listings}
\usepackage{adjustbox} 
\usepackage{multirow}
\usepackage{tabularx}
\usepackage{booktabs}
\usepackage{subcaption}
\usepackage{authblk}
\begin{document}

\title{OmniSCS: Omni Safety-Critical Scenario Synthesis \\ for Autonomous Driving via a Fully Editable Driving World}

\author{
    Xiaoyun Dong\textsuperscript{1}, 
    Qian Xu\textsuperscript{1}, 
    Yang Lu\textsuperscript{1}, 
    Yang Lou\textsuperscript{1}, 
    Yung-Hui Li\textsuperscript{2}, 
    and Jianping Wang\textsuperscript{1}
    }
\affil[1]{City University of Hong Kong } \affil[2]{Hon Hai Research Institute}

\twocolumn[{
\renewcommand\twocolumn[1][]{#1}
\maketitle
\begin{center}
\centering
\includegraphics[width=0.8\linewidth]{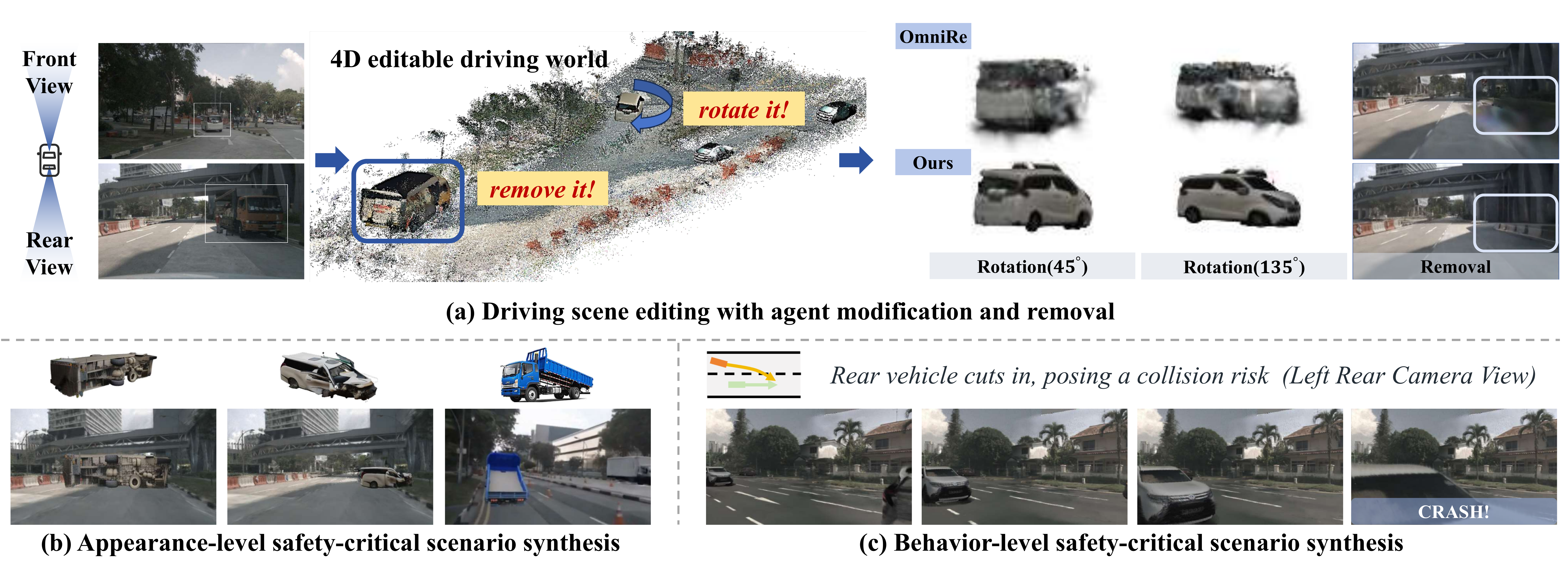}
\captionof{figure}{OmniSCS enables a fully editable driving world, supporting high‑fidelity appearance‑level and behavior‑level SCS synthesis for closed-loop testing. Please visit our project page (\href{https://omniscs.github.io/}{\texttt{https://omniscs.github.io/}}) for more results.}
\label{fig:primary_img}
\end{center}
}]

\begin{abstract}
The synthesis of safety-critical scenarios (SCS) and their evaluation through closed-loop simulations are crucial for developing robust autonomous driving systems. A key aspect of this process involves editing agent states in both appearance and trajectory levels within existing scenes. However, current methods struggle to preserve data fidelity after scene editing and fail to efficiently generate high-quality SCS through such modifications. To overcome these limitations, we propose OmniSCS,  an innovative system that generates photorealistic SCS with high physical fidelity while enabling closed-loop testing in synthetic environments.
OmniSCS comprises two key modules: 1) A Fully Editable Driving World Construction module that maintains high-fidelity agent appearance and background during scene editing via dual-strategy agent reconstruction and depth-refinement background reconstruction methods.
2) A SCS Synthesis module that facilitates object insertion and agent trajectory editing to synthesize diverse SCS while preserving data fidelity. Experiments on nuScenes, Waymo, and KITTI datasets show that OmniSCS outperforms state-of-the-art methods in edited scene fidelity. We further validate its ability to enhance autonomous driving algorithms and support real-time (13Hz) closed-loop testing. Overall, OmniSCS provides a safer, more effective, and cost-efficient solution for SCS optimization and testing in autonomous driving.
\end{abstract}

\section{Introduction}

Although recent advancements in modular and end-to-end (E2E) frameworks have enabled autonomous vehicles (AVs) to perform reliably in most routine driving scenarios~\cite{li2022bevformer,hu2023_uniad}, they still struggle in safety-critical scenarios (SCS). These high-risk situations—which represent only 1\% of driving data \cite{waymo-scs}—are difficult to address due to data scarcity and the inherent dangers of real-world closed-loop testing. As a result, high-fidelity data synthesis of safety-critical scenarios for effective closed-loop testing methods remains a crucial and yet underexplored area of research.

Following the classification in this field \cite{cc_survey,coda}, SCS are categorized into two levels: 1) Appearance-level, i.e., rare and unannotated objects, like excavators, forklifts, etc., common objects appearing in an unusual way or locations, like a truck overturned on the ground, resulting in failures in both perception module and E2E algorithms; 2) Behavior-level, i.e., trajectory collision risk between the ego vehicle and surrounding vehicles, resulting in motion planning failures in both planning module and E2E algorithms.

Recent research has explored reconstructing 3D scenes with high visual and geometric fidelity from recorded driving videos using Neural Radiance Fields (NeRFs) \cite{nerf} and Gaussian Splatting (GS) \cite{kerbl3Dgaussians, yang2023unisim, yan2024streetgs, chen2025omnire}. However, these approaches primarily reproduce data patterns from driving logs with similar camera perspectives, limiting generalization to unseen viewpoints. When agent trajectories deviate significantly from driving logs (Fig. \ref{fig:primary_img}(a)), artifacts such as blurring and ghosting degrade visual quality and hinder subsequent closed-loop evaluation. Although generative model-based approaches \cite{dm-vista,  dm-magicdrive, dm-drivedreamer2, dm-drivearena} are capable of synthesizing diverse driving scenarios, the inherently stochastic nature of diffusion processes causes temporal incoherence in the generated sequences, especially when editing the agent’s trajectories during behavior-level SCS synthesis and closed-loop testing. 
To the best of our knowledge, no existing work can synthesize high-fidelity SCS
while maintaining data quality after appearance and behavior editing, nor support reliable closed-loop testing in such synthetic scenarios. To address this unresolved problem, the key challenges are summarized below:

\textbf{Challenge 1: Maintaining Fidelity During Scene Editing.}
When modifying agent trajectories (e.g., lane changes, left/right turns), the agents' appearance must remain consistent with their driving behavior. However, existing scenario data lacks sufficient detail to reconstruct agents with high fidelity and motion continuity during movement. This challenge is especially acute for agents under low-resolution conditions or occlusion, where irreversible information loss further complicates accurate reconstruction.
Additionally, moving static objects can create a significant background deficiency in occluded regions. Such discrepancies in agent appearance and background continuity can lead to incorrect detections, rendering the scenario ineffective.

\textbf{Challenge 2: Effective Synthesis of Safety-Critical Scenarios.}
Although the fully editable driving world can synthesize both appearance-level and behavior-level SCS, data quality remains a significant challenge. For appearance-level SCS synthesis, the cross-modal discrepancy between 2D object images and 4D driving world models hinders the integration of objects into the background.
For behavior-level SCS synthesis, simply editing agents based on risk-inducing trajectories from prior SCS generation works \cite{scs-strive, scs-lctgen} causes noticeable artifacts in the corresponding rendered sensor data.

To address these challenges, we propose OmniSCS, a novel high-fidelity SCS synthesis framework based on real-world driving logs, illustrated in Fig~\ref{fig:primary_img}. The primary novelty of our approach lies in pioneering the simultaneous achievement of two critical objectives: 1) generating photorealistic SCS with high physical fidelity, and 2) enabling closed-loop testing within these synthetic environments. The framework consists of two key modules: 1) the Fulling Editable Driving World Construction (EDW) module, which tackles challenge 1 by reconstructing consistent $360^{\circ}$ appearances for agents and hole-free background representations. 2) the Safety-Critical Scenario Synthesis module, which addresses challenge 2 by facilitating the insertion of appearance-level objects into suitable locations and editing agents’ trajectories to generate behavior-level risky maneuvers. This enables 1-to-N SCS synthesis, allowing the generation of diverse scenarios from a single original scene, thereby enhancing AVs performance and supporting real-time closed-loop testing in dynamic environments through sensor feedback. 

OmniSCS offers two significant advantages. First, it can render photorealistic images,
outperforming state-of-the-art methods with FID improvements of 11.5\% on NuScenes, 14.5\% on KITTI, and 17.36\% on Waymo. Furthermore, 3D detection algorithm tested on OmniSCS-synthesized data achieves the highest mAP (+4.6\%) among image synthesis methods, verifying the data’s superior realism for downstream AVs tasks. Second, augmenting the training data with OmniSCS-synthesized SCS leads to significant improvements in both perception accuracy (+5.3\% mAP) and end-to-end driving safety (9.7\% lower collision rate). OmniSCS also supports real-time 13Hz rendering of large-scale driving scenes on an NVIDIA RTX 3090 GPU. Building on this capability, we integrate OmniSCS into a closed-loop simulator \cite{dm-drivearena}, showing that OmniSCS-synthesized SCS enables efficient AVs performance evaluation.

\section{Related Work}
\subsection{High-fidelity Data Generation}
High-fidelity data for autonomous driving can be produced via two approaches: reconstruction from real-world data and generation using learned models. Recent advances in neural rendering, particularly Neural Radiance Fields (NeRF) and 3D Gaussian Splatting (3DGS), have enabled photorealistic scene reconstruction for autonomous driving applications~\cite{yan2024streetgs, chen2025omnire, zhouHUGSIMRealtimePhotorealistic2024, chenSNeRFAutonomousDriving2024, wu2023mars}. These techniques excel at synthesizing novel viewpoints and reconstructing static environments with high fidelity. However, when agent behaviors significantly deviate from the captured data, these methods can produce blurring and ghosting artifacts, limiting their effectiveness in synthesizing SCS involving agent behavior changes. Rather than reconstructing real-world scenes, another type of approach is to generate synthetic driving scenarios directly. Recent advances in diffusion models~\cite{dm-vista, dm-magicdrive, dm-drivearena, dm-dd4d, dm-driveforge, dm-drivewm} offer the ability to generate controllable and diverse synthetic video data. These models accept multimodal inputs, such as camera images, Bird’s Eye View (BEV) layouts, and text prompts, to generate highly customizable driving scenarios. However, current diffusion models struggle to control the fine-grained appearance of individual objects in dynamic scenes, limiting their effectiveness for appearance-level SCS synthesis.

\subsection{Safety-critical Scenario Generation}
Traditional SCS generation approaches~\cite{scs-strive, scs-lctgen, scs-advsim} primarily focus on synthesizing adversarial behaviors of road agents to evaluate the planning performance of autonomous driving systems. These methods perturb agent trajectories while maintaining physical plausibility, thereby creating collision-prone interactions that challenge planning algorithms. However, these approaches typically emphasize behavior-level simulation and lack the generation of corresponding photorealistic sensor data aligned with agent behaviors, rendering them incompatible with perception or E2E autonomous driving systems that rely on sensor inputs such as camera images and LiDAR point clouds.

\section{Method}
\label{method}
\begin{figure*}[!t]
\centering
\includegraphics[width=0.9\textwidth]{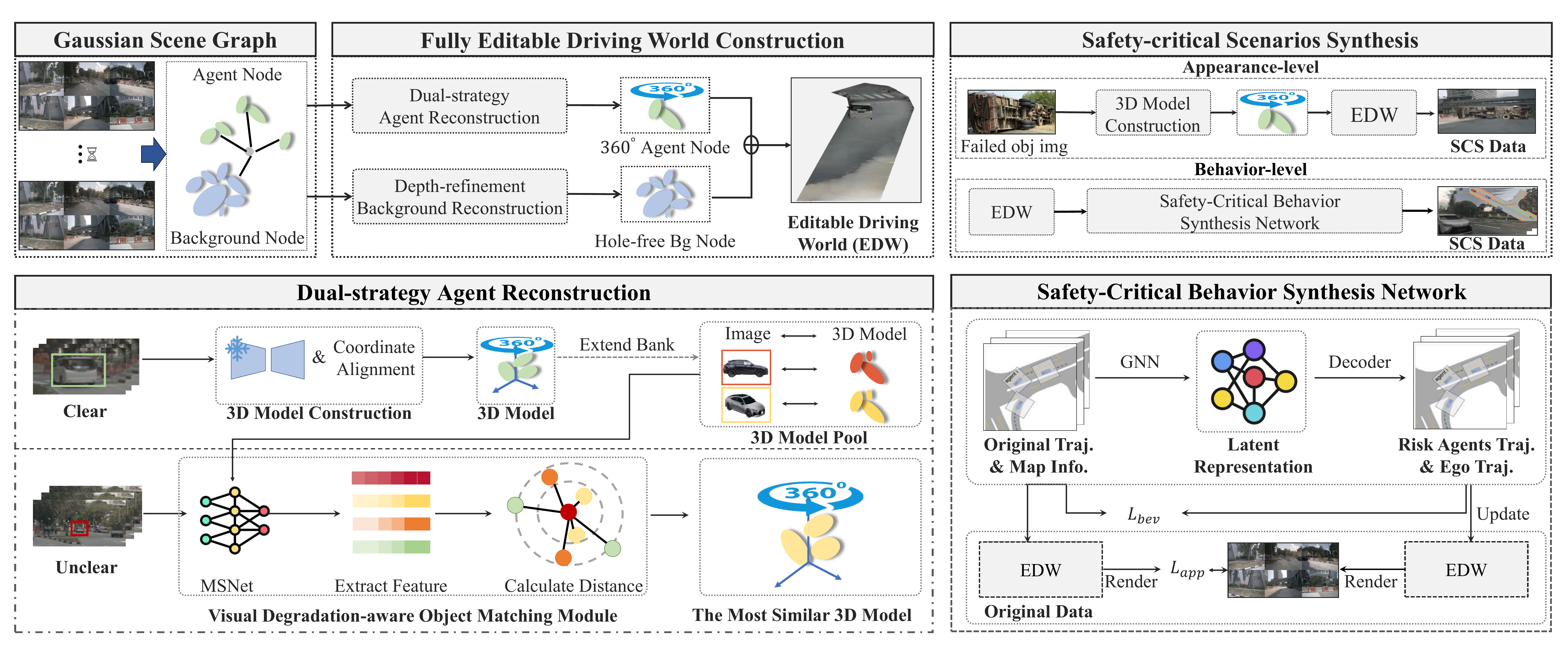}
\caption{The framework of OmniSCS. From driving videos, we build a Gaussian scene graph to model dynamic environments. We synthesize 360$^{\circ}$ agent nodes via dual-strategy reconstruction and optimize the occluded background via depth-refinement background reconstruction to construct a fully editable driving world. This facilitates appearance-level and behavior-level SCS synthesis. We then use the joint adversarial losses on trajectories and appearances to generate high-fidelity data.
}
\label{fig:arch-ar}
\vspace{-0.3cm}
\end{figure*}

We propose OmniSCS, a high-fidelity SCS synthesis framework depicted in Fig.~\ref{fig:arch-ar}. First, the Gaussian-Scene-Graph-based Fully Editable Driving World (EDW) Construction (Sec. \ref{sec:method-dw}) enables dynamic control over the movement, addition, and removal of all agents, allowing AVs to interact with others in the driving scene and supporting closed-loop evaluation. Second, the Safety-Critical Scenario Synthesis (Sec. \ref{sec:method-scs}) enables scalable, cost-effective, high-fidelity SCS synthesis at both appearance and behavior levels.

\subsection{Fully Editable Driving World Construction}
\label{sec:method-dw}
This module tackles the issue of significant artifacts caused by direct scene editing (outlined in Challenge 1). As shown in Fig.~\ref{fig:arch-ar}, OmniSCS first takes a Gaussian scene graph as input and 
decomposes the scene graph into separate agent nodes and a background node. Next, the dual-strategy agent reconstruction module reconstructs consistent $360^{\circ}$ appearances for all agent nodes. Simultaneously, the depth-refinement background reconstruction module eliminates holes—particularly those behind large static objects—within the background node, producing a complete and seamless representation. By merging these refined components, we can obtain a fully editable driving world.

\subsubsection{Gaussian Scene Graph for Driving Scenarios}

Real-world driving scenarios typically include dynamic road agents, such as moving vehicles. To enable the separation and subsequent editing of these agents, we adopt a Neural Scene Graph (NSG)~\cite{yan2024streetgs, chen2025omnire, zhouHUGSIMRealtimePhotorealistic2024}, which represents driving scenes as structured graphs learned from real-world sensor and ego-pose data, to
decompose the scene into background node \( G_{bg} \) and agent node \( G_A \). Each Gaussian blob within these nodes is parameterized by a set of properties, including position \( \mu \), rotation \( q \), scale \( s \), opacity \( \alpha \), and spherical harmonic coefficients \( c \). Notably, the agent node \( G_A \) comprises: 1) a local representation \( G_{A-loc} \), with Gaussian blobs positioned in the agent’s local coordinate system; 2) a rigid transformation matrix \( T_A \in SE(3) \), which encodes a rotation matrix \( R \) and a translation vector \( T \).  This maps the agent’s local coordinate system to the global world coordinate system, enabling the computation of the agent’s representation in the global coordinate through the operation:  
\begin{equation}
G_{A} = T_{A} \otimes G_{A-\text{loc}} = \left( R\mu + T, \ \text{Rot}(R, q), \ s, \ c, \ \alpha \right)
\label{eq:GA}
\end{equation}
Here, \( \text{Rot}(R, q) \) denotes the rotation of the quaternion \( q \) via the rotation matrix \( R \). Based on this formulation, agent behaviors can be easily modified by adjusting $T_{A}$.

\subsubsection{Self-improving Dual-strategy Agent Reconstruction}
\label{sec:method-ar}
As shown in Fig.~\ref{fig:arch-ar}, this module reconstructs 3D models with $360^{\circ}$ appearance and geometry for both clear and degraded agents in driving videos. Generative models can effectively reconstruct 3D representations from clear objects, but inferring novel-view appearance and geometry from limited observations of degraded objects remains a classic ill-posed problem. To address it, we propose a self-improving dual-strategy approach. First, a scalable 3D model construction pipeline builds an extensive model pool from clear objects, which can be continuously extended. Second, a degradation-aware matching module aligns degraded objects with appropriate models in the pool, enhancing reconstruction accuracy and robustness despite visual impairments.

\textbf{Scalable 3D Model Construction.} For agents with clear appearances, we use Trellis~\cite{trellis} to generate 3D Gaussian representations \(G\). A coordinate calibration module transforms \(G\) into the standard vehicle coordinate system, with the \(+x\) axis aligned with the vehicle’s heading. To standardize different vehicle types, we convert Gaussian blob centers into point clouds and align them with predefined templates using the Iterative Closest Point algorithm \cite{chetverikov2002trimmed}, selecting the best registration to calibrate the pose. Based on this approach, we have constructed a comprehensive 3D model pool that covers common vehicle types, providing detailed 3D representations of dynamic driving agents. Our comprehensive model pool includes 3D Gaussian models and 30 keyframe images (sampled at $12^\circ$ angular resolution)  for 440 popular vehicles across categories like sedans, SUVs, and trucks. As more scenes are processed, additional high-quality models can expand the pool’s coverage.

\textbf{Visual Degradation-aware Object Matching Module.}
Our aim is to accurately associate visually degraded target objects in driving logs, denoted as \(\text{ImgA}\), with high-resolution reference objects from the 3D model pool \(\text{ImgP}\). Once matched, the corresponding 3D Gaussian model from \(\text{ImgP}\) replaces the degraded agent node in the NSG for editability. Robust matching under dynamic conditions, including scale variations, illumination shifts, occlusions, and viewpoints, poses significant challenges for traditional feature-matching methods (e.g., SIFT~\cite{lowe1999object}, HOG~\cite{dalal2005histograms}) and even pre-trained deep features~\cite{match-clip, match-dino,  match-seg}, often leading to inaccurate associations. The main reason is that when capturing an object at different scales or viewpoints, the network needs to encode varying features to produce precise detection bounding boxes or segmentation outputs for the same object across varying conditions.
To address this, we propose a multi-scale feature extraction network (MSNet) and object-matching method. MSNet can minimize Euclidean distances between features of visually similar object pairs, accommodating variation in scale, occlusion, illumination, or viewpoint. This inherent capability empowers our method to accurately identify the most visually similar reference object in \(\text{ImgP}\) for each degraded object in \(\text{ImgA}\). 

MSNet utilizes a pre-trained backbone \cite{osnet} to extract object-level features from both \(\text{ImgA}\) and \(\text{ImgP}\), and then trains the network based on the loss defined in Eq.\ref{eq:lom}. Specifically, at the training stage, two objects \(A\) and \(B\) are randomly selected from the object pool. For object \(A\), two key images are randomly sampled:  one as the query (\(\text{ImgP}_{a}^{query}\)), degraded via operations like downsampling or illumination adjustment to simulate driving log challenges, and the other as the positive (\(\text{ImgP}_{a}^{pos}\)). For object \(B\), a random image serves as the negative (\(\text{ImgP}_{b}^{neg}\)). These images are fed into MSNet to obtain corresponding feature embeddings \(f_a\), \(f_{pos}\), and \(f_{neg}\).  We define the object matching loss $L_{om}$ to ensure that features of similar objects remain close in Euclidean space while dissimilar objects are pushed farther apart. $L_{om}$ is formulated as:
\begin{equation}
\small \frac{1}{|P|} \sum_{a \in P} \max \left(\max _{\substack{a \in P}}\left\|f_a-f_{neg}\right\| - \min _{\substack{a \in P }}\left\|f_a-f_{pos}\right\| +\alpha, 0\right)    
\label{eq:lom}
\end{equation}
During inference, we accelerate the matching process by precomputing and storing feature embeddings of all object pool images using MSNet, denoted as \(f_P\). For each degraded object in the driving log, the region is cropped to form \(\text{ImgA}\), its feature \(f_a\) is extracted via MSNet, and the Hungarian algorithm identifies the closest match in \(f_P\). The matched 3D Gaussian model then replaces the local representation \( G_{A-loc} \) in the NSG, enabling full editability of degraded objects in the scene. 
 

\subsubsection{Depth-refinement Background Reconstruction}

Maintaining background consistency is crucial when moving large static objects (e.g., trucks, buses, or shipping containers). To accurately integrate the inpainted regions into the Gaussian background node using an inpainting model ~\cite{controlnet}, absolute depth information is required. However, depth estimation models~\cite{depthanything} generally provide only relative depth, such as inter-pixel relationships, rather than true absolute depth values.
To address these challenges, we propose a dedicated module that generates both plausible inpainted appearances and precise geometric information for occluded regions. Additionally, we introduce an efficient training strategy to seamlessly integrate the information into  \(G_{bg}\).

\textbf{Depth Refinement Network.} 
To ensure visual consistency, we design the depth refinement module based on the principle that the inpainted image should preserve the relationship between the relative depth estimated from the original image and the corresponding ground-truth absolute depth values. 
As outlined in Tab.~\ref{tab:depth_net_arch}, we introduce DRNet, a depth refinement model that takes the relative depth map~\cite{depthanything} from the original image ($\text{Depth}_{rel}$) as input and predicts the absolute depth map $\text{Depth}_{abs}$. DRNet is trained with a mean squared error (MSE) loss:
\begin{equation}
\mathcal{L}_{\text{MSE}} = \frac{1}{N} \sum_{i=1}^{N} \Big(Depth_{abs}^{(i)} - \psi(pos(G))^{(i)}\Big)^2
\end{equation}
Here, $\psi(pos(G))$ represents absolute depth, derived by projecting the position of the Gaussian blob from the NSG onto the image plane and computing the Euclidean distance to the ego vehicle. Due to projection sparsity, the loss is computed only over pixels with valid measurements, where $N$ is the number of such pixels. If LiDAR data is available, a similar procedure can be applied using the LiDAR point cloud. During inference, $\text{Depth}_{rel}$ from the inpainted image, producing accurate $\text{Depth}_{abs}$ for the inpainted regions. 
\begin{table}[t]
\small
\setlength{\tabcolsep}{2pt}
\centering
\caption{Architecture of the depth refinement network.}
\begin{tabular}{lccc}
\hline
Layer             & Kernel & Output Channels & Activation \\
\hline
Conv2d            & 3     & 32             & LReLU + BN  \\
Conv2d            & 3     & 64             & LReLU + BN  \\
Spatial Attention~\cite{woo2018cbam} & 7     & 64    & Sigmoid         \\
Conv2d            & 3     & 32             & LReLU + BN  \\
Conv2d            & 3     & 1              & -            \\
\hline
\end{tabular}
\label{tab:depth_net_arch}
\vspace{-0.5cm}
\end{table}

\textbf{RoI-guided Background Node Training.} 
This method enhances NSG training efficiency by selectively optimizing only relevant regions of large static objects in the background node set $G_{bg}$.
For each large static object, we expand its 3D bounding box $(x, y, z, w, h, l, \text{rot})$ by a fixed scaling factor to define a Region of Interest (RoI), ensuring smooth transitions between inpainted areas and the original image. Gaussian blobs in $G_{bg}$ with positions $\mu$ inside this RoI are updated during training, while those outside remain frozen to minimize computation.
Finally, we re-optimize the Gaussian blobs within the RoI using the inpainted results and refined depth information, yielding a photorealistic rendering and geometric consistency with the updated scene.

\subsection{Safety-Critical Scenario Synthesis}
\label{sec:method-scs}
Building upon the fully editable driving world, we synthesize appearance-level and behavior-level safety-critical scenarios, respectively.



\subsubsection{Appearance-level SCS Synthesis}
We aim to achieve appearance-level SCS synthesis by compositing the target into the driving world, thus enabling 1-to-N appearance-level SCS generation. However, the cross-modal discrepancy between the 2D image of the target and the 4D driving world model introduces a significant representational gap, increasing the complexity of the synthesis task. 
To address this challenge, we design a synthesis pipeline that achieves high-fidelity in both the visual appearance and geometric placement of objects within the 4D driving environment.

Since appearance-level SCS are 
algorithm-specific, we first identify appearance-level objects $A_{scs}$ that fail to be detected using standard metrics~\cite{metric-det, metric-mota}. 
For each identified object, we crop its associated 2D images from the driving logs. Leveraging these 2D images and our presented scalable 3D model construction method, we obtain axis-aligned 3D object models, denoted as \(G_{ascs-loc}\). These models are then composited into a diverse driving world, enabling the generation of realistic and editable appearance-level SCS.
By replacing the \(G_{A-loc}\) of any agent node in EDW with \(G_{ascs-loc}\), the misdetected object can be composited into the 4D driving world. This ensures that \(A_{scs}\) inherits the original agent’s driving behavior. 
User-specified poses \(T_{a_{scs}}\) (e.g., overturned vehicles) can also be applied using Eq.~\ref{eq:GA}, allowing flexible placement of synthesized objects within a scene and across diverse 4D driving environments. 
By compositing objects with specified poses, our method generates synthesized scenes with automatic ground-truth annotations, greatly reducing the cost of appearance-level SCS data collection and labeling.

\subsubsection{Behavior-level SCS Synthesis} 
The primary challenge of behavior-level SCS synthesis lies in maintaining both agents and background fidelity when modifying agents’ trajectories from their original paths. When the ego vehicle follows significantly altered adversarial trajectories, the surrounding vehicles and background environment must correspondingly adapt to these new viewpoints, resulting in notable artifacts in the rendered sensor data. 
To this end, we propose joint adversarial losses on trajectories and appearances (as illustrated in Fig.~\ref{fig:arch-ar}), enhancing STRIVE~\cite{scs-strive} by adding appearance fidelity constraints. This enables the synthesis of diverse, high-quality behavior-level SCS data with consistent visual coherence despite trajectory modifications.




A graph neural network is utilized to capture the distribution of historical traffic data, including trajectories of surrounding agents and the ego vehicle. A decoding module then predicts future trajectories of all agents $\hat{T}_{t}$ and the ego vehicle $\hat{T}^{plan}_{t}$ within the scene. 
To balance scenario risk and data plausibility, we introduce a joint optimization loss $L_\text{adv}$ (Eq.~\ref{eq:scs_loss}) that perturbs agent trajectories in latent space while ensuring kinematic feasibility and fidelity to sensor data:
\vspace{-2pt}
\begin{equation}
    {L}_\text{adv} = \mathcal{L}_\text{{beh}} + \mathcal{L}_\text{{app}} + \mathcal{L}_{\text{restore}}
    \label{eq:scs_loss}
\end{equation}
\vspace{-2pt}
The adversarial behavior loss ${L}_\text{beh}$ (Eq. \ref{eq:adv-beh}) encourages controlled agents to minimize the positional distance between their trajectories and the planner’s trajectory:
\vspace{-2pt}
\begin{align}
\mathcal{L}_\text{beh} = \sum_{i=1}^N \sum_{t=1}^T \delta_t^i \left\lVert \hat{T}_t^i - \hat{T}_t^{\text{plan}} \right\rVert^2 
\label{eq:adv-beh}
\end{align}
\vspace{-2pt}
Here, the weighting factor $\delta_t^i$ gives higher importance to agents whose trajectories are closer to the planner’s to encourage adversarial interactions:
$\delta_t^i = \frac{
  \exp\left(-\left\lVert \hat{T}_t^i - \hat{T}_t^{\text{plan}} \right\rVert\right)
}{
  \sum_{j=1}^N \sum_{t=1}^T \exp\left(-\left\lVert \hat{T}_t^j - \hat{T}_t^{\text{plan}} \right\rVert\right)
}$.

The adversarial appearance loss $\mathcal{L}_\text{app}$ (Eq.~\ref{eq:adv-app}) is designed to ensure that the data generated from perturbed trajectories do not exceed the sparse viewpoints available in the original driving logs. This constraint enables the reliable generation of high-quality sensor data, even for novel trajectories not directly observed during data collection.
To enforce this, both the original and perturbed trajectories are rendered into image space using a differentiable rendering function $Re$, and perceptual features are extracted using a CNN $F$. This loss minimizes the discrepancy between the features extracted from the original and perturbed scene representations, thereby constraining appearance changes and preserving fidelity in the synthesized sensor outputs:
\begin{align}
\small
\mathcal{L}_\text{app}=\sum_{t=1}^T\left\|F\left(Re\left(\hat{NSG}\right)\right)-F\left(Re\left(NSG\right)\right)\right\|^2
\label{eq:adv-app}
\end{align}
Here, $\hat{NSG}$ and $NSG$ represent the perturbed and original Gaussian Scene Graphs, respectively.
Finally, the restoration loss $\mathcal{L}_\text{restore}$ constrains the perturbed latent trajectories to remain close to their initial states, thereby preserving the plausibility of the synthesized data while ensuring stability during optimization.

\section{Experiments}
\label{ex}
In this section, we compare our model with state-of-the-art data synthesis methods to evaluate the quality of generated data during scene editing.
We then demonstrate that our synthesized SCS data can effectively improve the performance of autonomous driving systems and enable closed-loop system testing.
Finally, we present ablation studies for each module to verify their effectiveness.
\subsection{Data Synthesis Quality}
\label{sec:data-quality}

We evaluate the realism of edited scenes on two primary dimensions: photorealism and algorithmic data fidelity.

\subsubsection{Photorealism}
We select challenging scenes from the NuScenes~\cite{dataset-nus}, Waymo~\cite{dataset-waymo}, and KITTI~\cite{dataset-kitti} datasets, covering heavy traffic, extreme weather, and lighting scenarios. Detailed test dataset information is available on our project page. As the ground-truth sensor data is unavailable after editing (due to object translation or rotation), we evaluate photorealism using the Fréchet Inception Distance (FID).

Quantitative results in Tab.~\ref{tab:data-quality-nus}(a) and Tab.~\ref{tab:data-quality-waymo-kitti} show that OmniSCS significantly outperforms baselines in scene editing, achieving efficient real-time rendering at 10 Hz, 13 Hz, and 15 Hz on an NVIDIA RTX 3090 GPU for these datasets, respectively. This highlights its practicality for closed-loop simulations and evaluations. 
Visualizations in Fig.~\ref{fig:editing} illustrate that baselines like OmniRE~\cite{chen2025omnire} and StreetGS~\cite{yan2024streetgs} suffer from blurring issues, while generative methods, including WorldDreamer~\cite{dm-drivearena} and DreamForge~\cite{dm-driveforge}, fail to maintain scene consistency. Moreover, DeformableGS~\cite{deformgs} struggles to effectively reconstruct dynamic vehicles. In contrast, OmniSCS achieves superior consistency and visual fidelity, demonstrating the effectiveness and robustness of our editable driving world reconstruction.

\twocolumn[{
    \renewcommand\twocolumn[1][]{#1}
    \begin{@twocolumnfalse} 
        \centering
        \small
        \captionof{table}{Quantitative evaluation of scene editing on Nuscenes. Best results are in bold, second best are underlined.}
        \begin{tabularx}{0.9\linewidth}{Xcccccccccc}
        \toprule
        & \multicolumn{5}{c}{\textbf{(a) Photorealism}} & \multicolumn{5}{c}{\textbf{(b) Data fidelity for Algorithms}} \\
        Method & \multicolumn{2}{c}{Translation (FID)$\downarrow$} & \multicolumn{3}{c}{Rotation (FID)$\downarrow$} & \multicolumn{2}{c}{Translation (mAP)$\uparrow$} & \multicolumn{3}{c}{Rotation (mAP)$\uparrow$} \\
         & 2.0m & 3.5m & 90$^\circ$ & 180$^\circ$ & 270$^\circ$ & 2.0m & 3.5m & 90$^\circ$ & 180$^\circ$ & 270$^\circ$ \\
        \midrule
        StreetGS & \underline{62.42} & \underline{67.26} & \underline{75.45} & 63.59 & 77.23 & 36.4 & 25.3 & 15.7 & 33.9 & 32.8 \\
        Omnire & 62.97 & 67.71 & 75.88 & 64.35 & \underline{76.03} & 36.8 & 24.8 & 30.5 & 34.3 & 30.5 \\
        PVG & 100.70 & 108.57 & 112.61 & 102.07 & 112.35 & 31.8 & 25.3 & 5.4 & 21.6 & 23.9 \\
        DeformableGS & 101.81 & 107.85 & 117.58 & 109.05 & 118.52 & 33.7 & 31.2 & 24.7 & 15.7 & 20.8 \\
        \midrule
        WorldDreamer & 75.75 & 79.86 & 87.40 & 71.51 & 86.42 & 40.9 & 31.2 & 38.4 & 38.3 & 38.5 \\
        DreamForge & 68.60 & 78.36 & 78.58 & \underline{59.21} & 79.61 & \textbf{45.8} & \underline{33.2} & \underline{41.1} & \underline{44.3} & \underline{41.0} \\
        \midrule
         Ours & \textbf{59.48 } & \textbf{63.37} & \textbf{65.18} & \textbf{52.51} & \textbf{65.54} & \underline{44.0} & \textbf{34.7} & \textbf{46.3} & \textbf{51.8} & \textbf{47.2} \\
        \bottomrule
        \end{tabularx}
        \label{tab:data-quality-nus}

        \small
        \captionof{table}{Quantitative evaluation of scene editing performance on Waymo and KITTI.}
        \begin{tabularx}{0.90\linewidth}{Xcccccccccc}
        \toprule
        & \multicolumn{5}{c}{\textbf{Waymo}} & \multicolumn{5}{c}{\textbf{KITTI}} \\
        Method & \multicolumn{2}{c}{Translation (FID) $\downarrow$} & \multicolumn{3}{c}{Rotation (FID) $\downarrow$}
            & \multicolumn{2}{c}{Translation (FID) $\downarrow$} & \multicolumn{3}{c}{Rotation (FID) $\downarrow$} \\
         & 2.0m & 3.5m & 90$^\circ$ & 180$^\circ$ & 270$^\circ$
            & 2.0m & 3.5m & 90$^\circ$ & 180$^\circ$ & 270$^\circ$ \\
        \midrule
        StreetGS & \underline{54.11} & \underline{59.93} & 79.52 & 70.52 & 76.61 & \underline{85.61} & \underline{90.13} & 107.41 & 100.06 & 110.77 \\
        Omnire & 54.39 & 60.40 & 80.71 & \underline{68.96} & 78.22 & 86.74 & 91.01 & \underline{106.41} & \underline{95.70} & \underline{107.92} \\
        PVG & 75.61 & 75.51 & \underline{75.20} & 72.32 & \underline{76.23} & 113.10 & 122.35 & 114.90 & 103.53 & 115.32 \\
        DeformableGS & 73.90 & 74.41 & 75.37 & 72.19 & 76.90 & 127.59 & 127.47 & 123.83 & 110.43 & 127.26 \\
        \midrule
        Ours & \textbf{49.32} & \textbf{52.32} & \textbf{64.37} & \textbf{51.34} & \textbf{62.15} & \textbf{73.24} & \textbf{81.62} & \textbf{94.49} & \textbf{74.05} & \textbf{94.26} \\
        \bottomrule
        \end{tabularx}
        \label{tab:data-quality-waymo-kitti}
        \vspace{0.5em}
        \includegraphics[width=0.9\linewidth]{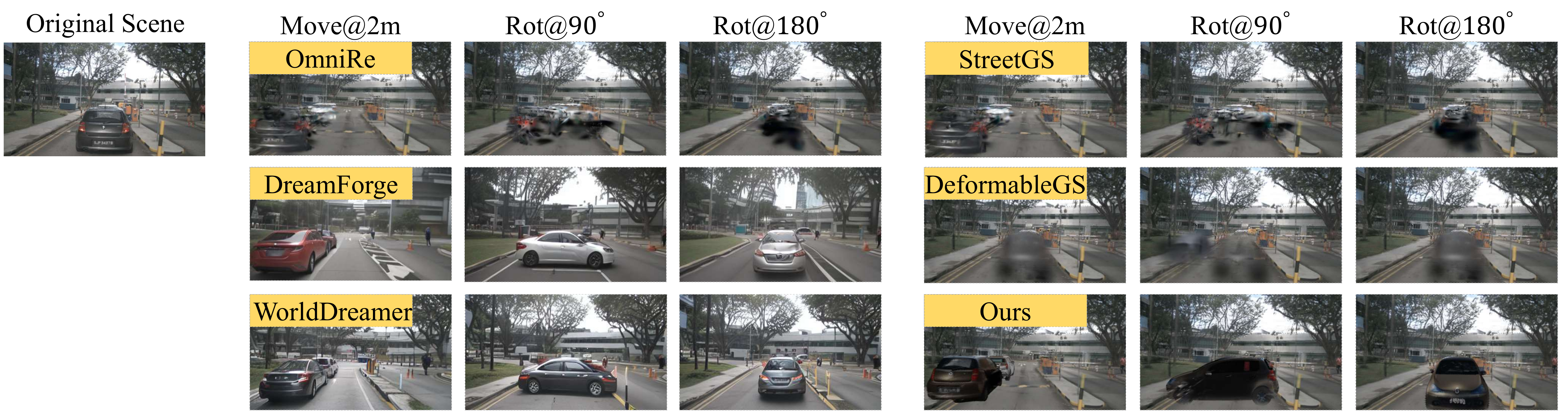}
        \captionof{figure}{Qualitative comparison of scene editing results under translation and rotation. Please visit our project page (https://omniscs.github.io/) for more results.
        } \label{fig:editing}
        \vspace{0.5em}
        \end{@twocolumnfalse}
}]
\noindent 
\subsubsection{Data Fidelity for Algorithms}
\label{sec:quality_algo}
Beyond photorealism, synthetic data for autonomous driving systems must also exhibit algorithmic realism. 
Since our goal is to enhance algorithm performance in synthesized SCS, we followed the setup of real-world corner case CODA dataset~\cite{coda} and selected 8 SCS for evaluation. We use UniAD’s 3D object detection module~\cite{hu2023_uniad} and assessed performance via Mean Average Precision (mAP). As shown in Tab.~\ref{tab:data-quality-nus}(b), our synthesized data yields significantly higher detection performance, achieving a $4.6\%$ improvement in mAP compared to the second-best method, DreamForge. In original scenarios, our method achieves the highest detection accuracy. Although DreamForge slightly outperforms at a 2.0 m translation, our approach shows a clear advantage under larger scene modifications, such as greater translations or rotations.
\subsection{Algorithm Performance Optimization and Closed-loop Testing in SCS}
\label{sec:opti}
\subsubsection{Algorithm Performance Optimization}
To validate the enhancement of model perception and driving safety in SCS using our synthesized data, we employ UniAD as the baseline, which can predict 3D detection and planning results. 
We use the same Nuscenes dataset described in the Data Fidelity for Algorithms section, randomly select 8 scenes from the NuScenes training set to generate both appearance-level and behavior-level SCS data. mAP evaluates the perception module, while L2 distance and collision rate assess the accuracy and safety of the planned trajectories.
Fig.~\ref{fig:ex3-pr} presents the mAP at different distance thresholds, showing that fine-tuning the perception module with OmniSCS-synthesized appearance-level SCS yields significant gains-outperforming the baseline by 5.3\% mAP at a 2.0 m threshold. 
In contrast, data synthesized by other methods can degrade detection performance. This highlights both the importance of SCS and the effectiveness of OmniSCS. 
Tab.~\ref{tab:e2e} further shows that behavior-level SCS data generated by OmniSCS significantly improves UniAD’s planning performance, enhancing overall driving safety. Specifically, the collision rate of UniAD after fine-tuning decreases by 9.7\% compared to the original UniAD. Fig.~\ref{fig:primary_img} presents the visualization results for appearance-level and behavior-level SCS synthesized by OmniSCS, respectively.
\subsubsection{Closed-loop Simulation}
\begin{figure}[!t]
\centering
\includegraphics[width=0.91\linewidth]{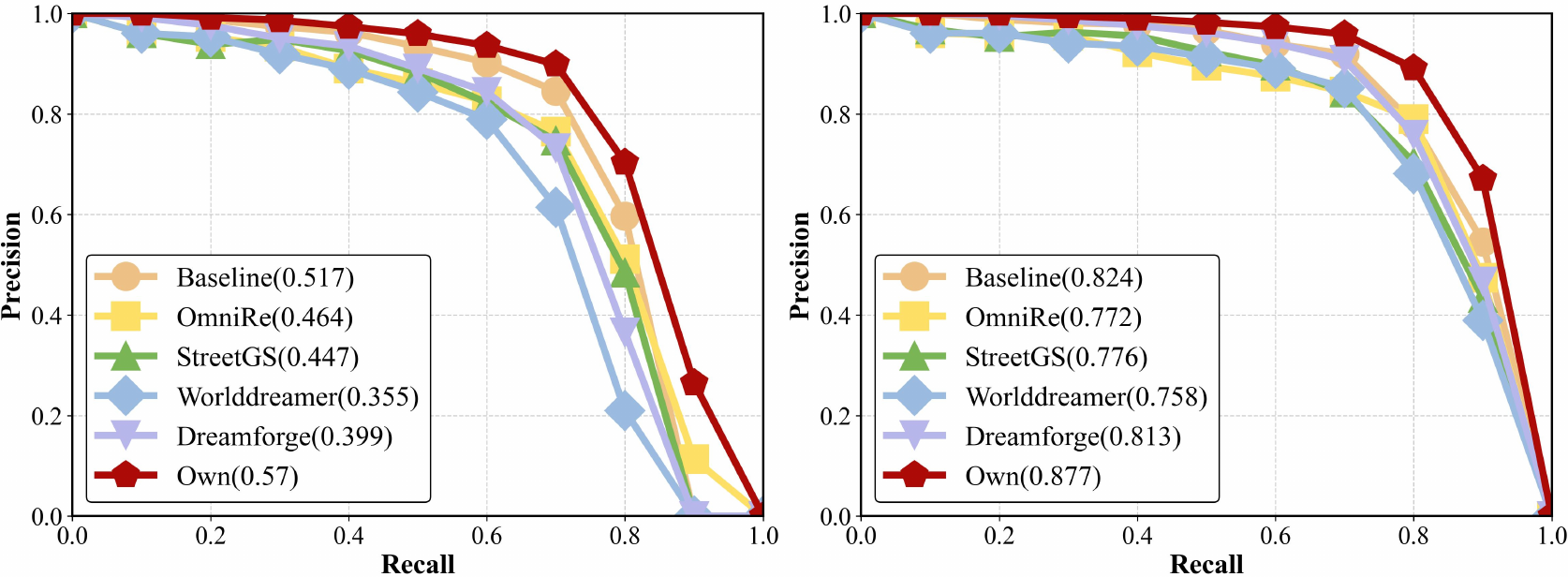}
\caption{Detection performance of UniAD perception module finetuned with different synthetic datasets under varying distance thresholds (Left: 2.0m, Right: 4.0m).}
\label{fig:ex3-pr}
\includegraphics[width=0.90\linewidth]{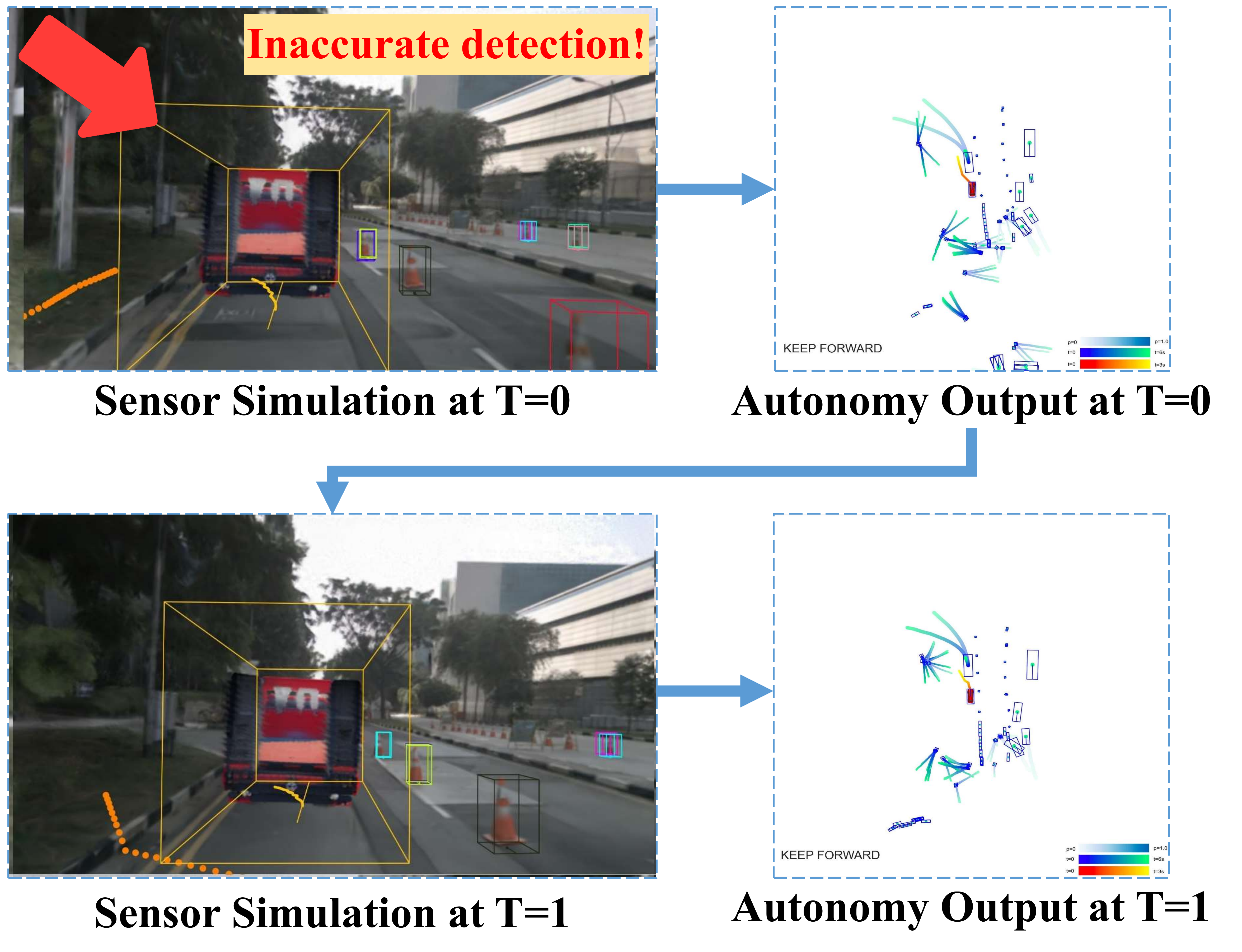}
\caption{Closed-loop evaluation in appearance-level SCS synthesized by OmniSCS. See our project page.}
\label{fig:closed-loop}
\includegraphics[width=0.90\linewidth]{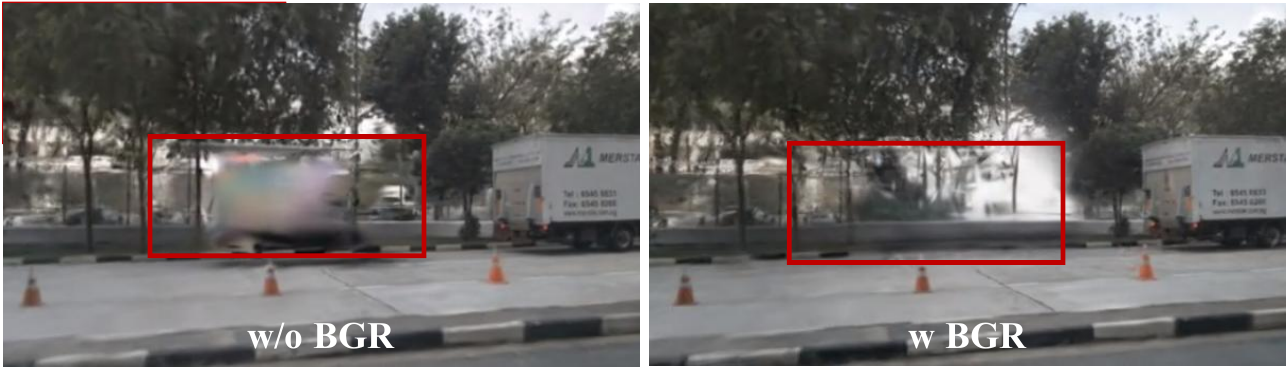}
\caption{Ablation of background reconstruction. 
}
\label{fig:abl-bg}
\vspace{-21pt}
\end{figure}

Fig.~\ref{fig:closed-loop} illustrates the closed-loop testing in OmniSCS-synthesized appearance-level SCS. When $T=0$, sensor data is rendered based on the current vehicle pose. UniAD processes this input to generate perception outputs and a planned trajectory. OmniSCS then renders new sensor data from the updated viewpoint, allowing UniAD to iteratively perform inference and planning. Additionally, this figure shows UniAD’s failure in detecting an unusual truck (resulting in erroneous detour trajectories), further validating OmniSCS’s ability to synthesize SCS.
\begin{table}[t]
\centering
\small
\setlength{\tabcolsep}{1mm}
\caption{Planning performance of UniAD finetuned with different synthetic datasets.}
\begin{tabularx}{0.95\linewidth}{Xcccccc}
\toprule
\multirow{2}{*}{Method} & \multirow{2}{*}{Collision Rate $\downarrow$} & \multicolumn{4}{c}{L2 (m) $\downarrow$} \\ 
 &  & 1.0s & 2.0s & 3.0s & Average \\ \midrule
Baseline & \underline{0.0145} & 1.17 & 1.83 & 2.38 & 1.79 \\ 
OmniRE & \underline{0.0145} & 1.11 & 1.78 & 2.36 & 1.75 \\
StreetGS & 0.0174 & 1.07 & 1.78 & 2.44 & 1.76 \\ 
WorldDreamer & 0.0203 & \underline{0.97} & \underline{1.57} & \underline{2.19} & \underline{1.58} \\
DreamForge & 0.0189 & 1.01 & 1.81 & 2.52 & 1.78 \\ \midrule
Ours & \textbf{0.0131 (9.7\% $\downarrow$)} & \textbf{0.73} & \textbf{1.44} & \textbf{2.14} & \textbf{1.44} \\ 
\bottomrule
\end{tabularx}
\label{tab:e2e}
\end{table}
\vspace{-1em}

\begin{table}[!t]
\centering
\small
\setlength{\tabcolsep}{1mm}
\caption{Ablation on dual-strategy agent reconstruction.}
\begin{tabularx}{0.95\linewidth}{XXcccc}
\toprule
\multirow{2}{*}{Online} & \multirow{2}{*}{Offline} & \multicolumn{1}{c}{Translation} & \multicolumn{2}{c}{Rotation} \\ 
 &  & FID @3.5m $\downarrow$ & FID @90$^\circ$ $\downarrow$ & FID @180$^\circ$ $\downarrow$ \\ \midrule
$\times$ & $\times$ & \underline{67.71} &75.88 & 64.35 \\
$\checkmark$ & $\times$ & 70.24 & \underline{66.81} & \underline{57.79}  \\
$\times$ & $\checkmark$ & 71.87 & 71.36 & 66.41\\
$\checkmark$ & \checkmark  & \textbf{63.37} & \textbf{65.18} & \textbf{52.51} \\ 
\bottomrule
\end{tabularx}
\label{tab:opti}
\end{table}

\begin{table}[!t]
\centering
\small
\setlength{\tabcolsep}{1mm}
\caption{Ablation study on appearance-level SCS.}
\begin{tabularx}{0.95\linewidth}{Xcccc}
\toprule
\multirow{2}{*}{Method}  & \multicolumn{4}{c}{mAP $\uparrow$ } \\ 
 &  1.0m & 2.0m & 3.0m & Average \\ \midrule
w/o Appearance-level SCS & \underline{0.446} & \underline{0.677} & \underline{0.791} & \underline{0.638} \\
w Appearance-level SCS & \textbf{0.570} & \textbf{0.778} & \textbf{0.877} & \textbf{0.742} \\ 
\bottomrule
\end{tabularx}
\label{tab:abl-app}
\end{table}

\begin{table}[!t]
\centering
\small
\setlength{\tabcolsep}{1mm}
\caption{Ablation study on behavior-level (Beh) SCS.}
\begin{tabularx}{0.95\linewidth}{Xccccc}
\toprule
\multirow{2}{*}{Method} & \multirow{2}{*}{Collision Rate $\downarrow$} & \multicolumn{4}{c}{L2 (m) $\downarrow$} \\ 
 &  & 1.0s & 2.0s & 3.0s & Average \\ \midrule
w/o Beh SCS & \underline{0.0174} & \underline{1.13} & \underline{1.94} & \underline{2.63} & \underline{1.79}\\ 
w Beh SCS & \textbf{0.0131} & \textbf{0.73} & \textbf{1.44} & \textbf{2.14} & \textbf{1.44}\\
\bottomrule
\end{tabularx}
\label{tab:abl-beh}
\vspace{-11pt}
\end{table}

\vspace{+7pt}
\subsection{Ablation Study}
\label{sec:abl}
\subsubsection{Dual-strategy Agent Reconstruction}
To evaluate the effectiveness of this module, we conduct ablation experiments by enabling the “online” synthesis strategy and “offline” object-matching strategy individually and jointly. The results (Tab.~\ref{tab:opti}) demonstrate that our method (combining both strategies) achieves the lowest FID scores under both translation and rotation editing, demonstrating superior performance compared to using either strategy alone.
\subsubsection{Depth-refinement Background Reconstruction} To evaluate the effectiveness of the background reconstruction, we conduct experiments on removing large static objects from the reconstructed world. The results show that our method preserves visual quality and produces visually realistic and coherent renderings without noticeable artifacts, as illustrated in Fig.~\ref{fig:abl-bg}. 
\subsubsection{Safety-Critical Scenario Synthesis}
To evaluate the effectiveness of SCS synthesis, we conduct experiments using both appearance-level and behavior-level SCS data. The results show that appearance-level SCS data significantly enhances UniAD’s perception module in terms of mAP across various distance thresholds (Tab.~\ref{tab:abl-app}), while behavior-level SCS data substantially improves UniAD’s planning
performance in terms of safety and accuracy (Tab.~\ref{tab:abl-beh}).

\section{Conclusion \& Discussion}
\label{con}

OmniSCS addresses two key challenges in synthesizing SCS and enabling closed-loop testing.  It enhances perception modules and end-to-end driving algorithms, improving the safety and robustness of autonomous driving. A limitation is that OmniSCS mainly focuses on rigid-body objects (e.g., vehicles and static obstacles), understating non-rigid entities like pedestrians and cyclists.
Future work will extend OmniSCS to incorporate these non-rigid objects’ appearance and behavioral complexities, enabling more comprehensive SCS synthesis and further performance gains.


\bibliographystyle{IEEEtran}
\bibliography{IEEEabrv,ref}

\end{document}